\documentclass[11pt]{article}
\usepackage{amsmath, amssymb, amsthm} 
\usepackage{subfigure} 
\usepackage{graphicx}
\usepackage{amssymb}
\usepackage{epstopdf}
\usepackage{booktabs}
\usepackage{multirow}
\title{A recommender system to restore images with impulse noise}
\author{Alfredo Nava-Tudela}
\date{January 19, 2017}
\begin{document}
\maketitle

\begin{abstract}
We build a collaborative filtering recommender system to restore images with impulse noise for which the noisy pixels have been previously identified. We define this recommender system in terms of a new color image representation using three matrices that depend on the noise-free pixels of the image to restore, and two parameters: $k$, the {\em number of features}; and $\lambda$, the {\em regularization factor}. We perform experiments on a well known image database to test our algorithm and we provide image quality statistics for the results obtained. We discuss the roles of {\em bias} and {\em variance} in the performance of our algorithm as determined by the values of $k$ and $\lambda$, and provide guidance on how to choose the values of these parameters. Finally, we discuss the possibility of using our collaborative filtering recommender system to perform image inpainting and super-resolution.
\end{abstract}

\section{Introduction} \label{sec: introduction}
Impulse noise in a grayscale image is characterized by the appearance of random single pixel errors on the image. This type of error occurs when there are malfunctioning pixel elements in the camera sensors, faulty memory locations, or timing errors in analog-to-digital conversion \cite{GonWoo2002,PlaVen2000}.

We shall model impulse noise in the following way. Let $a_{i,j}$ and $y_{i,j}$ be the pixel values at position $(i,j)$ in the original and noisy images, respectively. Let $[s_{\min}, s_{\max}]$ be the {\em dynamic range} of the image, i.e., the set of possible pixel values. Let $p$ be a number, called the {\em noise ratio}, such that $p \in [0,1]$, then
\begin{equation} \label{eq: impulse noise model}
y_{i,j} =
\begin{cases}
a_{i,j}, & \quad \text{with probability } 1-p \\
n_{i,j}, & \quad \text{with probability } p 
\end{cases}
\end{equation}
where $n_{i,j} \neq a_{i,j}$ and $n_{i,j} \in [s_{\min}, s_{\max}]$. If $n_{i,j} = s_{\min} \text{ or } s_{\max}$, we say that the image has {\em salt-and-pepper noise} \cite{Bovi2005}; and if $n_{i,j}$ is a (feasible) value chosen at random from  $\mathcal{U}(s_{\min}, s_{\max})$, we say that the image has {\em random-valued impulse noise} \cite{DonChaXu2007}.

In color images, impulse noise is modeled by considering its presence as defined above in any of its color components taken individually as grayscale images.

There are results to detect and correct impulse noise in images, see for example the various techniques described in \cite{DonChaXu2007,PlaVen2000}. Current approaches divide the task in two steps, firstly by identifying the likely pixels that have been corrupted by impulse noise, and secondly by applying a local filter to restore their values, see for example \cite{ChaHoNik2005,GarHueChuHe2005}. Here we will concern ourselves with the restoration of color images for which corrupted pixels have already been ``detected" by an impulse noise detection algorithm, say like ROLD \cite{DonChaXu2007}, a detection statistic for random-valued impulse noise. In particular, for our study, we will use the {\em TID2008} image database \cite{PonLukZelEgiCarBat2009}, which provides reference images and corresponding random-valued impulse noise counterpart examples. Unlike in the works referenced above, our goal will be to restore the noisy images with a {\em recommender system} which will substitute the pixels labeled as noise with a suggested value for each of them, as opposed to applying a local filter in each case.

This work is divided in the following way. In section \ref{sec: recommender system} we introduce recommender systems in general and the collaborative filtering system that we propose in particular. We then lay out the experimental methodology that we will use to test our collaborative filtering system in section \ref{sec: experimental setup} and report on the results obtained in section \ref{sec: results}. We follow with a discussion in section \ref{sec: discussion}, where we talk about the bias vs variance trade-off, see section \ref{sec: bias vs variance trade-off}; how to choose the two main parameters of our recommender system, the number of features $k$ and the regularization factor $\lambda$ that control that trade-off, see section \ref{sec: choice of k and lambda}; and a brief note on the new family of image decompositions that our recommender system is based upon and the implications for the system performance, see section \ref{sec: image decomposition}. Finally, we present our conclusions and future work directions in section \ref{sec: conclusions and future work}.

\section{A collaborative filtering recommender system} \label{sec: recommender system}
Recommender systems are algorithms and techniques that provide suggestions for their users on a given topic or class of objects. For example, a recommender system could suggest what items to buy, what music to hear, what movie to watch, or what news to read \cite{RicRokSha2011}. In our case, we are going to implement a recommender system that will suggest what value to give to a missing or corrupted pixel in an image, a novel approach.

The particular flavor of recommender system that we are going to use is a {\em collaborative filtering algorithm} \cite{RicRokSha2011}. In this modality, ``systems require recommendation seekers to express preferences by rating a dozen or two items, thus merging the roles of recommendation seeker and preference provider. These systems focus on algorithms for matching people based on their preferences and weighting the interests of people with similar tastes to produce a recommendation for the information seeker" \cite{Carr2002}. For example, consider a recommender system where the items are movies and the users rank them in a scale from 1 to 5. The recommender system would then compute a ranking for a movie that you have not yet ranked by taking into consideration the rankings given to that movie by other users and how similar your tastes are to the tastes of those users. Then, if the ranking is above certain threshold, the system would recommend that movie to you \cite{Ng2016}.

In our case, we are going to think of the rows in an image as the movies in the example above, and its columns as the users. Then our recommender collaborative filtering algorithm will provide a ranking for a missing pixel, which in this case will translate into its suggested value.

But why should a system like this work? Imagine that two people with very similar tastes in wine decide to rank 50 wines on a predetermined scale. Now suppose that the first person has scored all 50 wines and that the second one has done so for the first 49 wines in the list. If their rankings are very similar for the first 49 wines, we would expect that the second person would give a very similar ranking to the ranking that the first person gave to the 50th bottle of wine. That is intuitively why we expect the collaborative filtering algorithm to work.

\subsection{Collaborative filtering cost function} \label{sec: cost function}
Consider a grayscale image $\mathbf{Y}$ of size $m \times n$ with integer entries $y_{i,j} \in [0, 2^b - 1]$, where $b \in \mathbb{N}$ is the image's {\em bit depth}. Typically $b=8$. In this case $[0, 2^b - 1]$ is what we called the dynamic range of the image in section \ref{sec: introduction}. Consider a matrix $\mathbf{R}$ also of size $m \times n$ but with entries $r_{i,j} \in \{0, 1\}$, where $r_{i,j} = 1$ if we know the pixel value $y_{i,j}$ at position $(i,j)$ to be correct and 0 otherwise, i.e., the pixel value is missing or has been labeled as noise by our impulse noise detector. We will call such matrix $\mathbf{R}$ a {\em mask for} $\mathbf{Y}$.

Now imagine that for each row $i$ of $\mathbf{Y}$ we associate a vector $\mathbf{x}^{(i)} \in \mathbb{R}^k$, where $k \in \mathbb{N}$ is a fixed number of {\em features} that we are going to associate with that row. Similarly we associate a vector $\theta^{(j)} \in \mathbb{R}^k$ to column $j$ of $\mathbf{Y}$. Hence we will have $m$ vectors $\{\mathbf{x}^{(1)}, \mathbf{x}^{(2)}, \ldots, \mathbf{x}^{(m)}\} \subset \mathbb{R}^k$ and $n$ vectors $\{\theta^{(1)}, \theta^{(2)}, \ldots, \theta^{(n)}\} \subset \mathbb{R}^k$ associated with our image $\mathbf{Y}$. The collaborative filtering algorithm then considers these two sets of vectors and will model the prediction of the pixel value at position $(i,j)$ by computing $z_{i,j} = \mathbf{x}^{(i) \text{T}} \theta^{(j)}$, i.e., the inner product of $\mathbf{x}^{(i)}$ and $\theta^{(j)}$. In other words, if we form the matrices
\begin{equation} \label{eq: X and Theta}
\mathbf{X} =
\begin{bmatrix}
\mathbf{x}^{(1)} & \mathbf{x}^{(2)} & \cdots & \mathbf{x}^{(m)}
\end{bmatrix}
\quad \text{and} \quad
\Theta =
\begin{bmatrix}
\theta^{(1)} & \theta^{(2)} & \cdots & \theta^{(n)}
\end{bmatrix},
\end{equation}
then our {\em prediction (recommendation) matrix} is
\begin{equation} \label{eq: prediction matrix}
 \mathbf{Z} = \mathbf{X}^\text{T} \Theta.
 \end{equation}
Define the {\em collaborative filtering cost function for} $\mathbf{Y}$ as
\begin{equation} \label{eq: cost function}
J(\mathbf{X},\Theta) = J\left(\mathbf{x}^{(1)}, \ldots, \mathbf{x}^{(m)}, \theta^{(1)}, \ldots, \theta^{(n)}\right) =
\frac{1}{2} \sum_{(i,j)\, :\, r_{i,j} = 1} \left( \mathbf{x}^{(i)\text{T}}\theta^{(j)} - y_{i,j} \right)^2,
\end{equation}
where implicitly we have provided a mask $\mathbf{R}$ for $\mathbf{Y}$ as well.

To link matrices $\mathbf{X}$ and $\Theta$ meaningfully to $\mathbf{Y}$, we require that they solve the minimization problem
\begin{equation} \label{eq: minimization problem}
\min_{\mathbf{X},\Theta} J(\mathbf{X},\Theta).
\end{equation}
To see why this requirement is meaningful, let $\mathbf{R}$ be a mask for $\mathbf{Y}$ whose entries are all ones and assume that $J(\mathbf{X},\Theta) = 0$, then it is easy to see that $y_{i,j} = \mathbf{x}^{(i)\text{T}}\theta^{(j)}$ for all $(i,j)$. That is, $\mathbf{X}$ and $\Theta$ offer a perfect reconstruction (factorization) of image $\mathbf{Y}$. In other words $\mathbf{Y} = \mathbf{X}^\text{T}\Theta$.

\subsection{Regularization and normalization} \label{sec: regularization and normalization}
In section \ref{sec: cost function} we defined the collaborative cost function $J$ for an image $\mathbf{Y}$ that measures how good two matrices $\mathbf{X}$ and $\Theta$ approximate the known good (non noisy) values of the pixels of $\mathbf{Y}$ when the approximation is given by $\mathbf{X}^\text{T}\Theta$.

Two remarks are in order. Assume that $\mathbf{x}, \theta \in \mathbb{R}^k$ exactly approximate the pixel value $y$ at position $(i,j)$ in image $\mathbf{Y}$. In that case, $y = \mathbf{x}^\text{T} \theta$. Also assume that $k > 1$ and write the inner product of these two vectors as $\mathbf{x}^\text{T} \theta = d + x_k \theta_k$, where we focus our attention on the product of their last corresponding $k$-th vector components. Then,
\begin{equation}
y = d + x_k \theta_k\ \Leftrightarrow\ x_k \theta_k = c, \quad \text{where }
c = y - d.
\end{equation}
Without loss of generality, assume that $c \neq 0$. Then
\begin{align}
\theta_k = \frac{c}{x_k} \quad \text{and} \quad \theta_k^2 + x_k^2 &= \left( \frac{c}{x_k} \right)^2 + x_k^2, \nonumber \\
&= \frac{x_k^4 + c^2}{x_k^2}.
\end{align}
Observe that the real function of real variable $f(x) = \frac{x^4 + c^2}{x^2}$ has the following properties:
\begin{enumerate}
\item $\lim_{x \rightarrow \pm \infty} f(x) = +\infty,$ and $\lim_{x \rightarrow 0} f(x) = +\infty$.
\item $f$ has two minima. One at $x_1 = \sqrt{|c|}$ and another one at $x_2 = -\sqrt{|c|}$.
\end{enumerate}
The first remark then is that without any further restrictions, we can find an infinite number of vector pairs $(\mathbf{x}, \theta)$ for which their combined $\ell_2$-norm is arbitrarily large and for which $y = \mathbf{x}^\text{T}\theta$; and that there are at most two vector pairs $(\mathbf{x}, \theta)$ with minimal combined $\ell_2$-norm satisfying $y = \mathbf{x}^\text{T}\theta$. This prompts us to regularize the cost function $J$ defined in equation \eqref{eq: cost function} and propose instead the {\em regularized collaborative filtering cost function for} $\mathbf{Y}$ as
\begin{align} \label{eq: regularized cost function}
J_\lambda(\mathbf{X},\Theta) = J_\lambda\left(\mathbf{x}^{(1)}, \ldots, \mathbf{x}^{(m)}, \theta^{(1)}, \ldots, \theta^{(n)}\right) =&
\frac{1}{2} \sum_{(i,j)\, :\, r_{i,j} = 1} \left( \mathbf{x}^{(i)\text{T}}\theta^{(j)} - y_{i,j} \right)^2 + \nonumber \\
& \frac{\lambda}{2}\left( \sum_{i=1}^m \|\mathbf{x}^{(i)}\|_2^2 
+ \sum_{j=1}^n \|\theta^{(j)}\|_2^2 \right),
\end{align}
where $\lambda \geq 0$ is the {\em regularization factor}. Consequently, the minimization problem \eqref{eq: minimization problem} is equivalently transformed into the {\em regularized minimization problem for} $\mathbf{Y}$
\begin{equation} \label{eq: regularized minimization problem}
\min_{\mathbf{X},\Theta} J_\lambda(\mathbf{X},\Theta).
\end{equation}
The introduction of regularization reduces the search space for a solution of problem \eqref{eq: regularized minimization problem}, and also addresses the issue of {\em model overfitting}, of which we will talk in section \ref{sec: discussion}.

The second remark comes from the following observation. Assume that image $\mathbf{Y} \in \mathbb{R}^{m \times n}$ is not trivially zero and that a whole column $j$ of pixels is tagged as noise. If we set $\lambda > 0$, a solution to minimization problem \eqref{eq: regularized minimization problem} will necessarily return a matrix $\Theta$ for which $\theta^{(j)} = \mathbf{0}$, the zero vector. Then for any row $i$, the recommender system will return a value of $a = 0$ as the suggested value for the pixel at position $(i,j)$, given that $\mathbf{x}^{(i)\text{T}}\theta^{(j)} = 0$ in this case. This would be a very poor recommendation since at least there is a row $i$ for which the average value of the noise free pixels in that row is not zero, and assigning that average to the missing pixel would be a more meaningful suggestion than zero. To have an intuition as to why that is the case, think back to the scenario where users rank movies. For a new user that hasn't ranked any movies yet and wants to know what other users think of a particular movie, the recommender system would do a better job if it offered the average recommendation that other users have given to that movie than offering a ranking value of zero.

One way to address this problem would be to normalize the original image $\mathbf{Y}$ by subtracting from each row the average value of the noise-free pixel values in that row, and then solve problem \eqref{eq: regularized minimization problem} for this new matrix. For a noisy pixel at position $(i,j)$ we would then recommend  the value obtained by computing the corresponding inner product and adding to it the average value aforementioned. We describe this procedure in detail in the following section.

\begin{figure}[htbp]
\centering
	\subfigure[Matrix representation of image I23.]{\includegraphics[width=.993\textwidth]{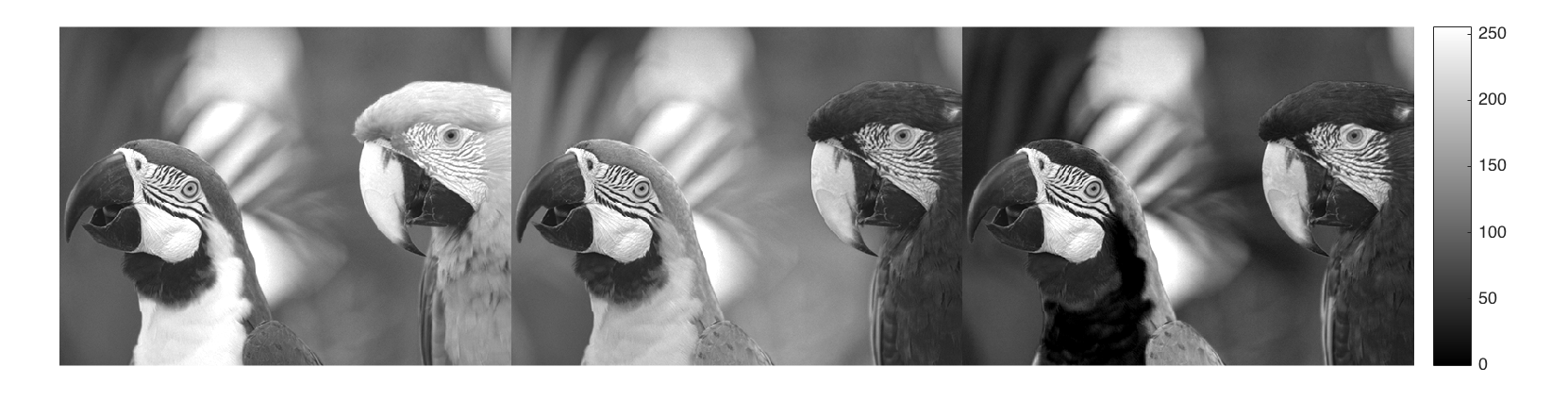} \label{fig: parrots}}
	\subfigure[Matrix representation of image I23\_06\_4.]{\includegraphics[width=.993\textwidth]{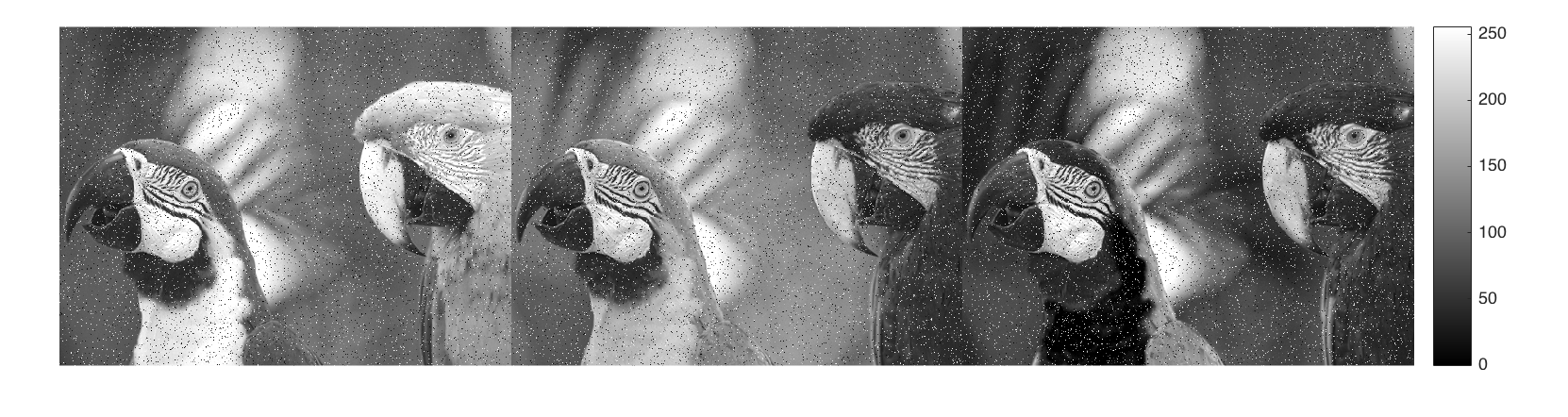} \label{fig: noisy parrots}}
\caption{Matrix representation of color images. The red, green, and blue channels are concatenated column wise from left to right.}
\label{fig: matrix representation of color images}
\end{figure}

\section{Experimental setup} \label{sec: experimental setup}
As mentioned in section \ref{sec: introduction}, we use the photographs in the {\em TID2008} image database \cite{PonLukZelEgiCarBat2009} to conduct our experiments. This database consists of 25 reference images, 17 types of distortions for each reference image, 4 different levels of each type of distortion. From these, we work with the images affected with random-valued impulse noise. The generative model for this distortion is described by equation \eqref{eq: impulse noise model}. The 4 levels of distortion correspond to a noise ratio of about $p, 2p, 4p$ and $8p$ for levels 1 through 4, respectively, with $p \approx 0.0085$.

Each color image in our database is a 24-bit depth color image of size $384 \times 512$ pixels which we manipulate into a single matrix of size $384 \times 1536$ where each of the 8-bit red, green, and blue color channels are concatenated column wise, see figure \ref{fig: matrix representation of color images} for examples. From this point onward, when we mention a color image we will think of it this way, unless we say otherwise, i.e., as a single matrix where all three color channels have been combined column wise.

With this treatment for color images in place, given a noisy image $\mathbf{Y}$ in our dataset, like the one depicted in figure \ref{fig: noisy parrots}, the experiment consists in following the steps described in the last paragraph of section \ref{sec: recommender system}. In particular, we have to first obtain a mask $\mathbf{R}$ for $\mathbf{Y}$ that identifies the pixels that have been corrupted with random-valued impulse noise. Here is where we can use, for example, the ROLD statistic \cite{DonChaXu2007} to mark pixels that are to be considered impulse noise.

\begin{figure}[htbp]
\centering
\includegraphics[width=\textwidth]{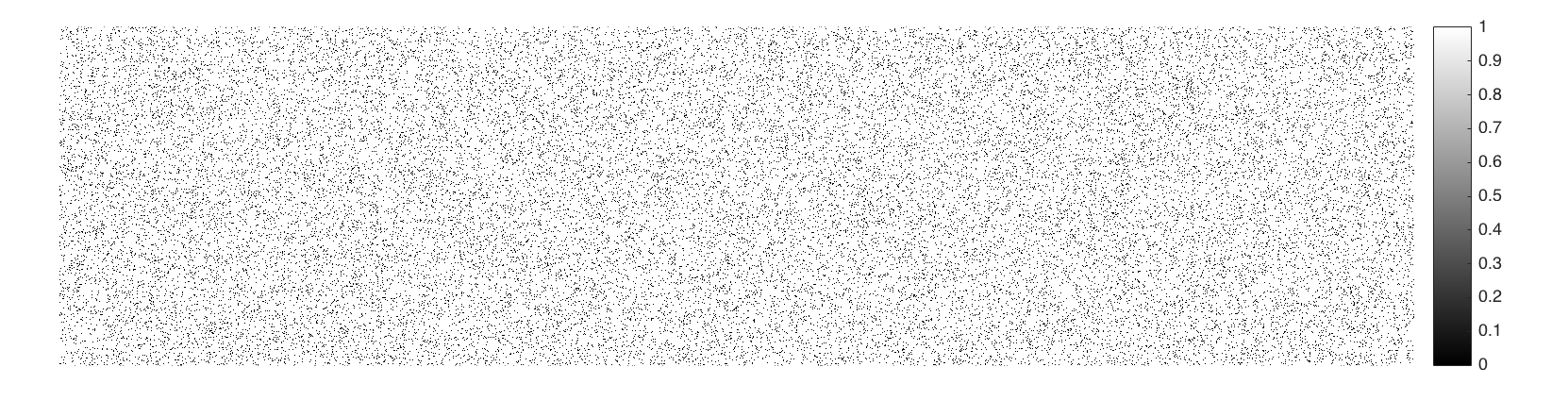}
\caption{Perfect detection mask $\mathbf{R}$ for image I23\_06\_4 given reference image I23. Black dots represent pixel locations where the images differ, and hence identify noisy pixels in I23\_06\_4.}
\label{fig: mask}
\end{figure}

The mask $\mathbf{R} = (r_{i,j}) \in \{0,1\}^{m \times 3n}$, with $m = 384$ and $n =  512$ in this case, is a matrix such that $r_{i,j} = 1$ if the pixel at position $(i,j)$ is considered noise-free, and zero otherwise, see figure \ref{fig: mask}.

In our case we count with the noise-free reference image $\mathbf{A}$, see figure \ref{fig: parrots}, and this means that we can obtain a perfect mask $\mathbf{R}$ for $\mathbf{Y}$ in the sense that it will identify correctly all pixels that are noise, and only those. This is not necessarily the case when we don't have a reference image that represents the grown truth. The relationships between the noise-free image $\mathbf{A}$, its related noisy image $\mathbf{Y}$, the mask $\mathbf{R}$ for $\mathbf{Y}$ (with respect to $\mathbf{A}$), and the {\em noise matrix} $\mathbf{N}$ is given by the following equation
\begin{equation} \label{eq: image decomposition}
\mathbf{Y} = \mathbf{A} \circ \mathbf{R} + \mathbf{N} \circ \neg \mathbf{R},
\end{equation}
where $\circ$ is the Hadamard product of matrices (entry-wise multiplication), and $\neg$ is the negation operator that changes a 1 for a 0 and vice versa. Note that the noise matrix $\mathbf{N}$ is assumed to satisfy the conditions of equation \eqref{eq: impulse noise model}, and it can be made unique if we impose the extra condition that it have the least number of nonzero entries.

Next we compute $\mathbf{B}$, the normalized version of the noisy image $\mathbf{Y}$, as described in the last paragraph of section \ref{sec: recommender system}, namely, let $\mu \in \mathbb{R}^m$ be the vector with entries $\mu_i = \frac{1}{\#\{j\, :\, r_{i,j} = 1\}} \sum_{j\, :\, r_{i,j}=1} y_{i,j}$, then define $\mathbf{B} \in \mathbb{R}^{m \times 3n}$ to be the matrix with entries $b_{i,j} = y_{i,j} - \mu_i$ if $r_{i,j} = 1$, and 0 otherwise. We now proceed to obtain a solution to problem \eqref{eq: regularized minimization problem} for the normalized color image $\mathbf{B}$. For this, we have to select a regularization factor $\lambda$ as well. We chose $\lambda = 0$ and $\lambda = 10$ to compare non-regularized and regularized solutions.

Since this is a minimization problem, a method such as gradient descent or any flavor of conjugate gradient can be used. Either of these two approaches requires choosing an initial value for matrices $\mathbf{X} \in \mathbb{R}^{k \times m}$ and $\Theta \in \mathbb{R}^{k \times 3n}$, see equation \eqref{eq: X and Theta}, from which the minimization algorithm then iterates to pursue a minima of the regularized collaborative filtering cost function $J_\lambda(\mathbf{X},\Theta)$, see equation \eqref{eq: regularized cost function}. We chose both matrices to have entries taken from the normal distribution with zero mean and variance one, $N(0,1)$, as initial values. The choice of the number of features $k$ will be discussed in section \ref{sec: discussion}. For our experiments we selected $k=1, 8, 308, 352, 380, 384$.

With this experimental setup, let $(\mathbf{X_{\mathbf{B},\mathbf{R},\lambda},\Theta_{\mathbf{B},\mathbf{R},\lambda}}) = {\arg \min}_{\mathbf{X},\Theta} J_\lambda(\mathbf{X},\Theta)$ be a solution pair for the regularized minimization problem for $\mathbf{B}$. We then propose image
\begin{equation} \label{eq: image restoration}
\mathbf{C} = \mathbf{Y} \circ \mathbf{R} + \left({\mathbf{X}_{\mathbf{B},\mathbf{R},\lambda}}^\text{T}\Theta_{\mathbf{B},\mathbf{R},\lambda} + \mu \cdot \mathbf{1}^\text{T}\right) \circ \neg \mathbf{R},
\end{equation}
as the reconstruction from the noisy image $\mathbf{Y}$, from mask $\mathbf{R}$ for $\mathbf{Y}$, and regularization factor $\lambda$, where $\mathbf{1} \in \mathbb{R}^{3n}$ is the vector with all ones, and $n = 512$ in this case. Recall that $\mu \in \mathbb{R}^m$ and hence $\mathbf{M} = \mu \cdot \mathbf{1}^\text{T} \in \mathbb{R}^{m \times 3n}$. Basically, if $\mathbf{R}$ is a perfect mask, as is the case in our experiments, we have that $\mathbf{Y} \circ \mathbf{R} = \mathbf{A} \circ \mathbf{R}$, and comparing equations \eqref{eq: image decomposition} and \eqref{eq: image restoration} gives that we are substituting the noise matrix $\mathbf{N}$ with the sum of the prediction matrix $\mathbf{Z}_{\mathbf{B},\mathbf{R},\lambda} = {\mathbf{X}_{\mathbf{B},\mathbf{R},\lambda}}^\text{T}\Theta_{\mathbf{B},\mathbf{R},\lambda}$, see equation \eqref{eq: prediction matrix}, with the normalization term $\mathbf{M}$, to restore image $\mathbf{A}$ from its noisy version $\mathbf{Y}$. Observe that, by design, $\|(\mathbf{A} - (\mathbf{Z}_{\mathbf{B},\mathbf{R},\lambda} + \mathbf{M})) \circ \mathbf{R}\|_2$ is small, which justifies why we would make that substitution of $\mathbf{N}$. We   can therefore call the matrix $\mathbf{D}_{\mathbf{B},\mathbf{R},\lambda} = \mathbf{Z}_{\mathbf{B},\mathbf{R},\lambda} + \mathbf{M} = {\mathbf{X}_{\mathbf{B},\mathbf{R},\lambda}}^\text{T}\Theta_{\mathbf{B},\mathbf{R},\lambda} + \mathbf{M}$ the {\em denoising matrix}.

\section{Results} \label{sec: results}
In this section we present the results that we obtain when we follow the experimental protocol, described in section \ref{sec: experimental setup}, for all of the reference photographs in the {\em TID2008} image database \cite{PonLukZelEgiCarBat2009} that are affected with each of the four impulse noise levels available therein. The noise levels are controlled by the noise ratio parameter $p$, see equation \eqref{eq: impulse noise model}, which has in this case, on average, the values of 0.0085, 0.0169, 0.0339, and 0.0678.

To measure the quality of the reconstructions we obtain, we utilize the {\em peak signal-to-noise ratio} (PSNR) \cite{Bovi2005, TauMar2002}, measured in decibels (dB), and the {\em mean structural similarity index} (MSSIM), a number from 0 to 1, where 1 corresponds to two identical images. This image quality statistic is designed to correlate with the way humans perceive image quality \cite{WanBovSheSim2004}.

\begin{figure}[htbp]
\centering
\includegraphics[width=\textwidth]{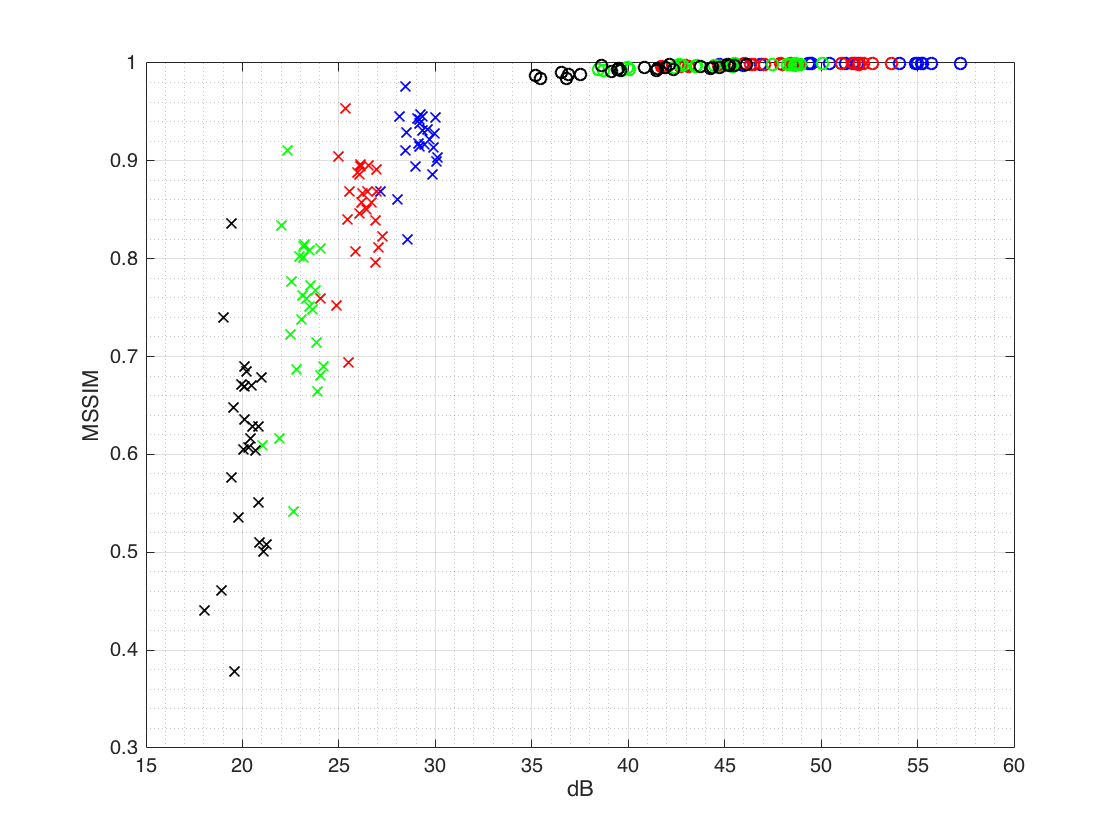}
\caption{Image quality metrics. Each ``x" represents a noisy image for which its peak signal-to-noise ratio in dB and mean structural similarity index (MSSIM) values have been plotted. The ``o" points correspond to their denoised reconstruction counterparts. The colors codify for the noise ratio $p$ used, see equation \eqref{eq: impulse noise model}: Blue, $p = 0.0085$; red, $p = 0.0169$; green, $p = 0.0339$;  and black, $p = 0.0678$. The reconstructions used $k = 352$ features and a regularization factor of $\lambda = 11$. In these experiments we used a perfect detection mask $\mathbf{R}$ for each image. See section \ref{sec: recommender system}.} \label{fig: image quality metrics}
\end{figure}

The first set of results we present can be seen in figure \ref{fig: image quality metrics}. For each image in the {\em TID2008} database affected with impulse noise we plot its PSNR and MSSIM image quality metrics. Then we process each image applying to it the experimental protocol described in section \ref{sec: experimental setup}, and plot on the same figure both image quality metrics for the reconstruction. We can observe that for all images and all four levels of noise ratio $p$ the improvements in quality are very satisfactory, especially as measured by MSSIM. The choice of the number of features $k=352$ and regularization factor $\lambda=11$ in our experiments is justified in section \ref{sec: discussion}.

\begin{figure}[htbp]
\centering
\includegraphics[width=\textwidth]{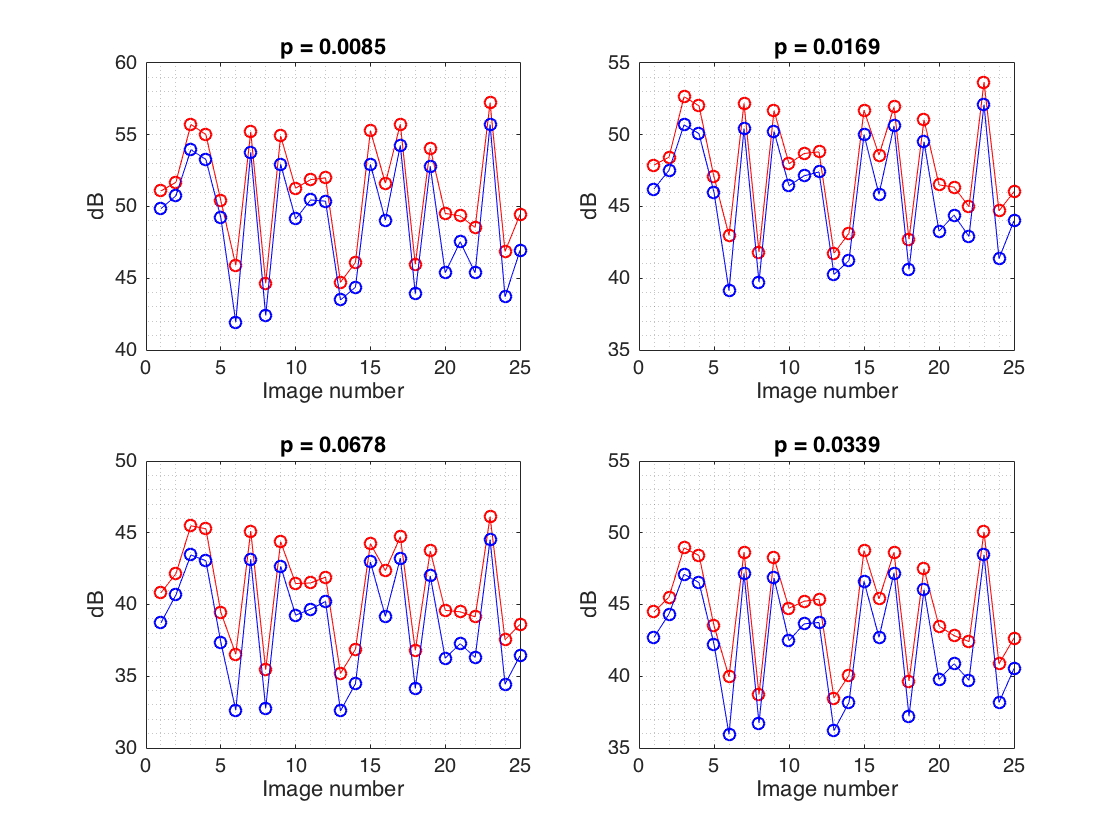}
\caption{Regularized vs non-regularized solutions comparison. For each of the four noise ratio values of $p$ used, increasing clockwise from top left, we plot the peak signal-to-noise ratio (dB) of the denoised images obtained with a regularization factor of $\lambda = 11$ and $k=352$ features (in red), and non-regularized solutions with $k=352$ features (in blue). All the regularized solutions have better PSNR than the non-regularized ones. In these experiments we used a perfect detection mask $\mathbf{R}$ for each image.} \label{fig: regularization effect on PSNR}
\end{figure}

\begin{figure}[htbp]
\centering
\includegraphics[width=\textwidth]{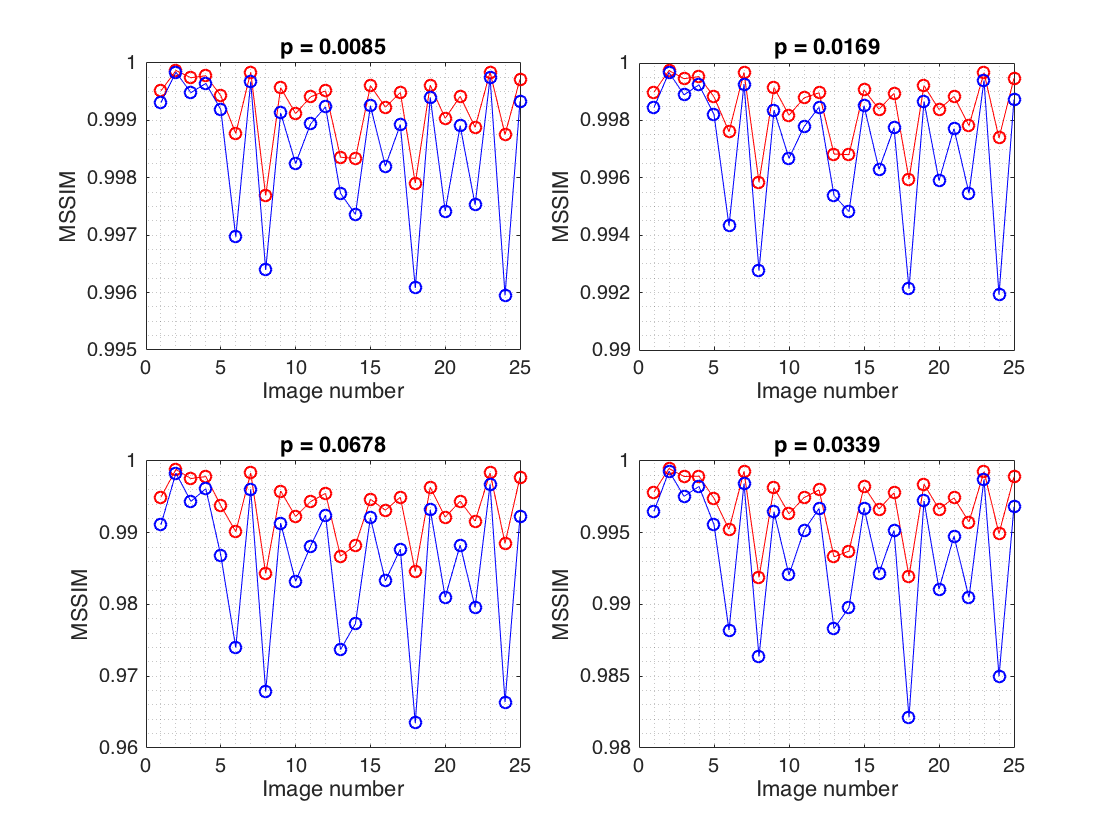}
\caption{Regularized vs non-regularized solutions comparison. For each of the four noise ratio values of $p$ used, increasing clockwise from top left, we plot the mean structural similarity index (MSSIM) of the denoised images obtained with a regularization factor of $\lambda = 11$ and $k=352$ features (in red), and non-regularized solutions with $k=352$ features (in blue). All the regularized solutions have better MSSIM than the non-regularized ones. In these experiments we used a perfect detection mask $\mathbf{R}$ for each image.} \label{fig: regularization effect on MSSIM}
\end{figure}

The next set of results that we present pertain to regularization, see sections \ref{sec: regularization and normalization} and \ref{sec: discussion} for definitions and justification of its use. Figures \ref{fig: regularization effect on PSNR} and \ref{fig: regularization effect on MSSIM} show the effects of regularization on PSNR and MSSIM for $k=352$ number of features and a regularization factor of $\lambda=11$. The choices of these two numbers is discussed in section \ref{sec: discussion}.

The results confirm that regularizing the solutions is a good idea that yields better reconstruction results, if marginally in some instances, but dramatic in others. The figures show the results for all 25 images contained in the {\em TID2008} database.

\begin{figure}[htbp]
\centering
	\subfigure[Image I07 (detail)]{\includegraphics[width=0.32\textwidth]{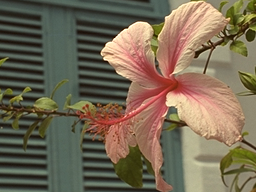}}
	\subfigure[Noisy]{\includegraphics[width=0.32\textwidth]{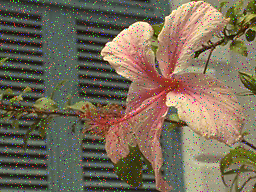}}
	\subfigure[Reconstruction]{\includegraphics[width=0.32\textwidth]{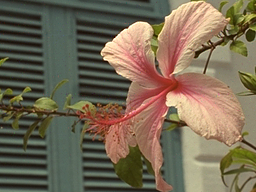}}
	\subfigure[Image I09]{\includegraphics[width=0.32\textwidth]{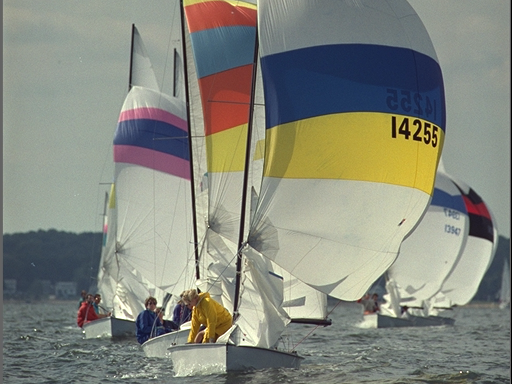}}
	\subfigure[Noisy]{\includegraphics[width=0.32\textwidth]{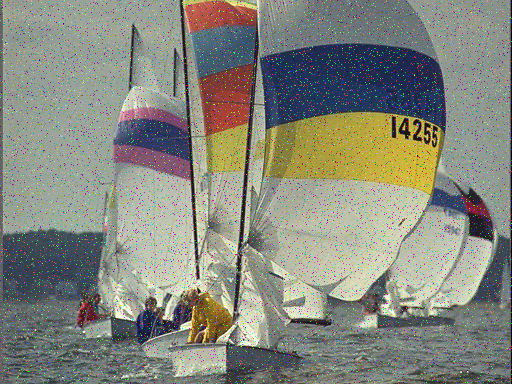}}
	\subfigure[Reconstruction]{\includegraphics[width=0.32\textwidth]{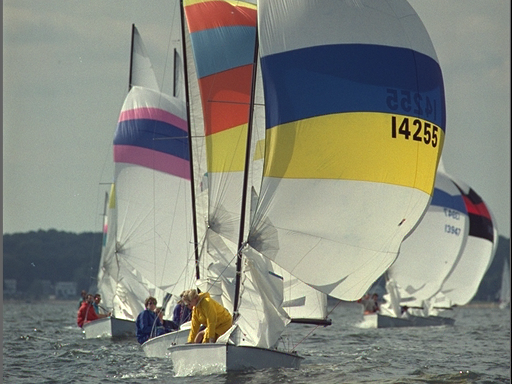}}
	\subfigure[Image I20]{\includegraphics[width=0.32\textwidth]{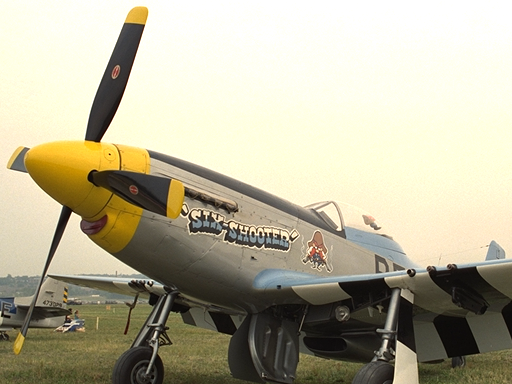}}
	\subfigure[Noisy]{\includegraphics[width=0.32\textwidth]{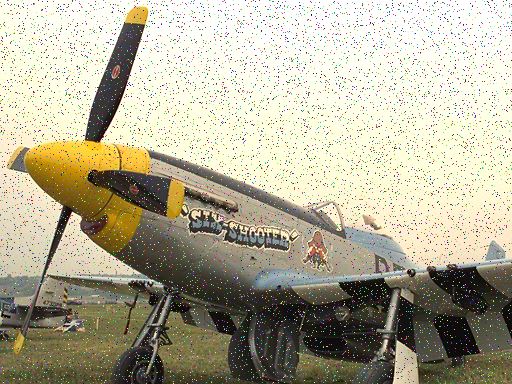}}
	\subfigure[Reconstruction]{\includegraphics[width=0.32\textwidth]{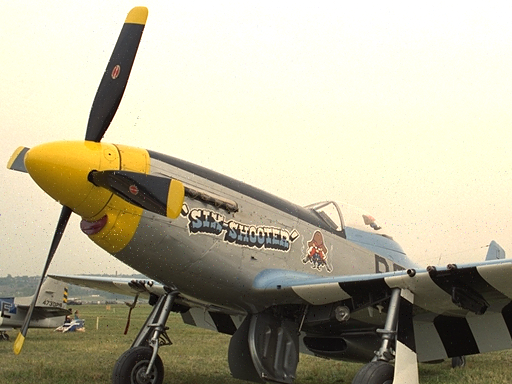}}
	\subfigure[Image I23]{\includegraphics[width=0.32\textwidth]{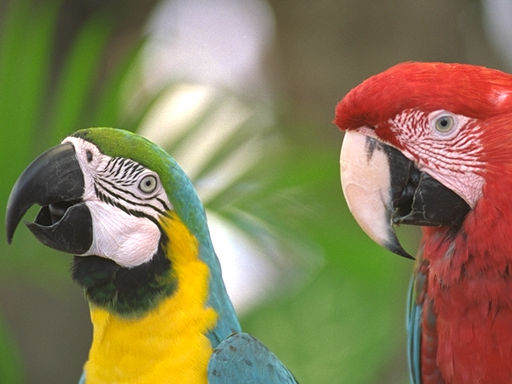}}
	\subfigure[Noisy]{\includegraphics[width=0.32\textwidth]{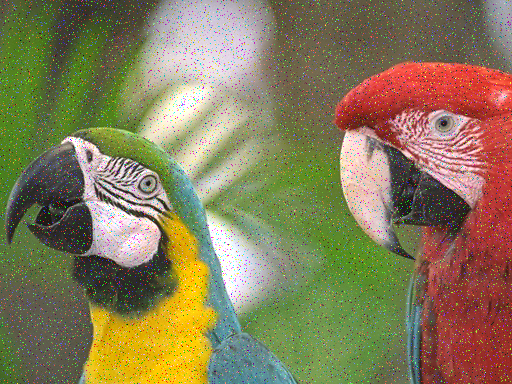}}
	\subfigure[Reconstruction]{\includegraphics[width=0.32\textwidth]{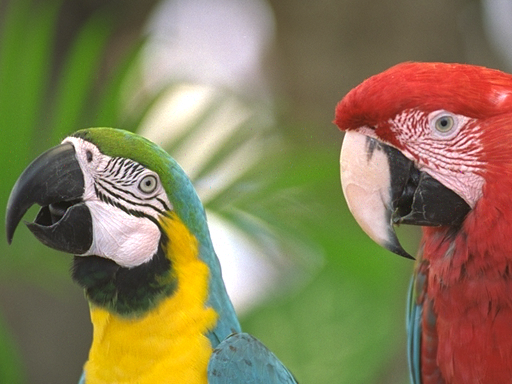}}
\caption{Images in their original, noisy ($p = 0.0678$), and filtered forms. The reconstructions were obtained using a regularization factor of $\lambda = 11$, $k = 352$ features, and a perfect detection mask $\mathbf{R}$.} \label{fig: image comparisons}
\end{figure}

Finally, we show in figure \ref{fig: image comparisons} a selection of images affected by random-valued impulse noise for a noise ratio of $p=0.0678$, their reconstruction with our methodology, and the originals from which the noisy versions were generated for comparison purposes.

\section{Discussion} \label{sec: discussion}
\subsection{Bias vs variance trade-off} \label{sec: bias vs variance trade-off}
When designing or choosing a model to represent a data set, we are interested in quantifying to what  degree the model appropriately captures its {\em essence}. To help us quantify the goodness of a model we will show how the concepts of {\em bias} and {\em variance} play a role.
\begin{figure}[htbp]
\centering
	\subfigure[]{\includegraphics[width=.75\textwidth]{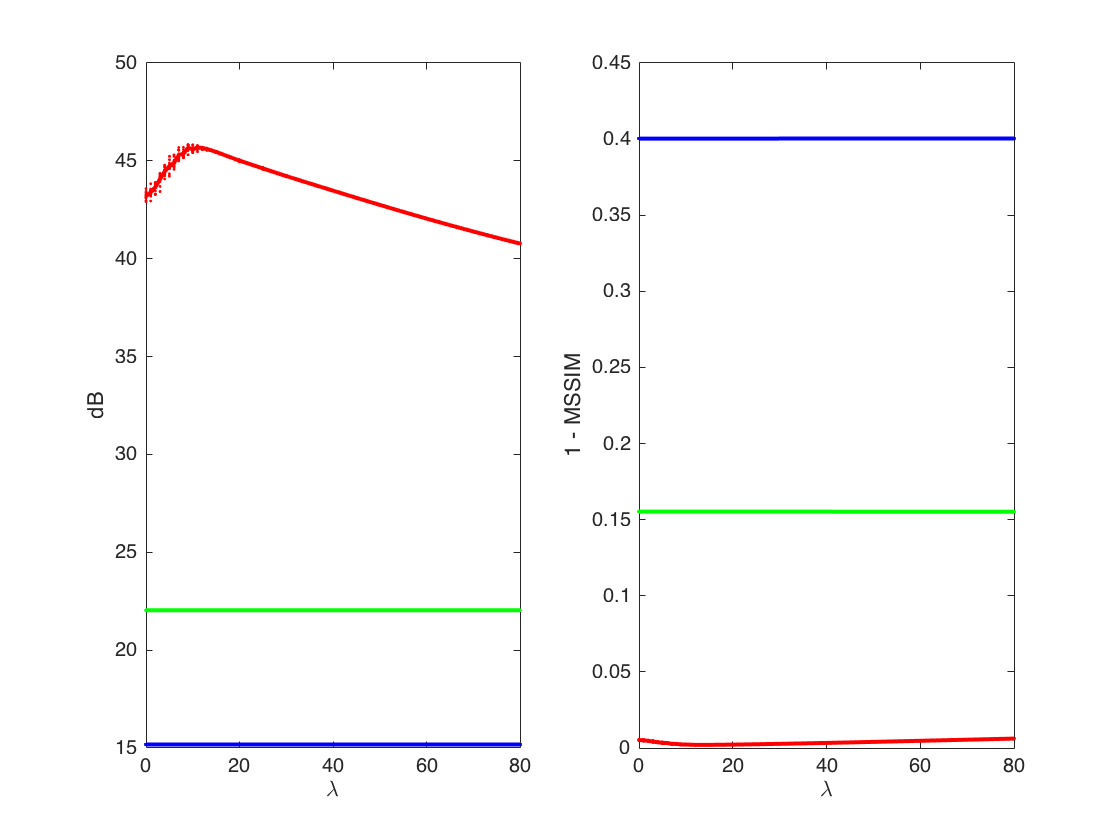} \label{fig: bias vs variance}}
	\subfigure[$k=1, \lambda=0$]{\includegraphics[width=.3\textwidth]{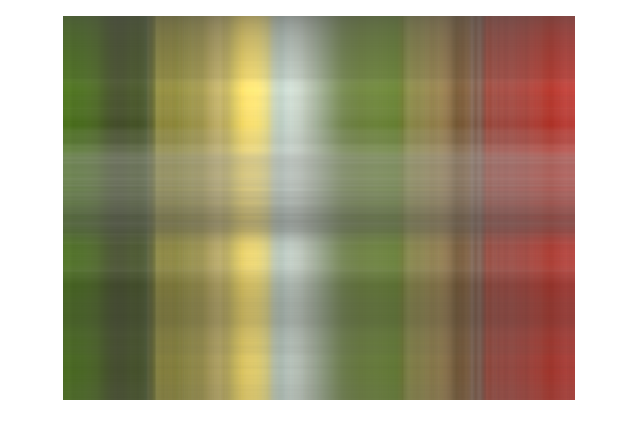} \label{fig: k=1}}
	\subfigure[$k=8, \lambda=0$]{\includegraphics[width=.3\textwidth]{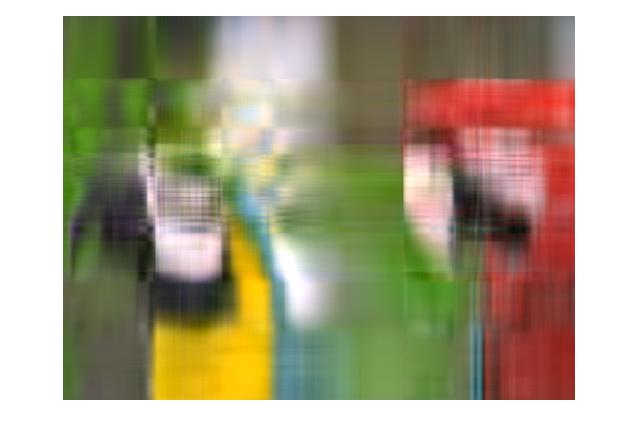} \label{fig: k=8}}
	\subfigure[$k=352, \lambda=0$]{\includegraphics[width=.3\textwidth]{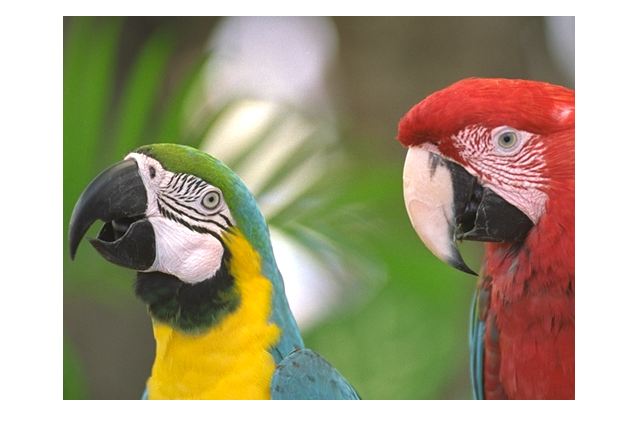} \label{fig: k=352}}
\caption{High bias vs high variance. A {\em high bias} model will fail to represent for any value of the regularization factor $\lambda$ the original image and offer a recommendation that has a very high error. See for example figures \subref{fig: k=1} and \subref{fig: k=8} above, corresponding to the blue and green lines in \subref{fig: bias vs variance}, respectively. A {\em high variance} model will represent with a small error the original, even for a regularization factor of $\lambda=0$. It will be optimal for a positive value of the regularization factor, as it tends to overfit the data given without regularization. See figure \subref{fig: k=352}, for example, which corresponds to the red line in \subref{fig: bias vs variance}.}  \label{fig: bias-variance trade-off}
\end{figure}

Suppose that we conduct the following experiment. Given a set $\{x_1, x_2, \ldots, x_n\} \subseteq X$ we associate to each $x_i$ a {\em sample value} $y_i \in \mathbb{R}$, forming a set of observations $\{(x_1,y_1), (x_2,y_2), \ldots, (x_n,y_n)\}$, which we call the {\em training set}. Assume that $y_i = f(x_i) + \epsilon$, where $\epsilon$ is a random variable with zero mean and variance $\sigma^2 > 0$, which we call {\em noise}, and $f : X \rightarrow \mathbb{R}$ is an unknown deterministic function which we wish to determine. We typically try to do this by {\em training} a function $\tilde{f} : X \rightarrow \mathbb{R}$ over the training set that minimizes the mean square error of the approximation by $\tilde{f}$ to the sample values $\{y_i\}$ in it. Ideally, we would like to come up with a function $\tilde{f}$ that would also minimize the mean square error for sample values of other possible points $x_0$ for which we haven't gotten any data yet, i.e., $x_0 \in X \setminus \{x_i\}$, so that we could make good predictions $\tilde{f}(x_0)$ with our model $\tilde{f}$. Essentially, we want to minimize the mean square error over $X$, which we decompose as follows
\begin{align} \label{eq: mean square error decomposition}
E\left[\big(y - \tilde{f}\big)^2\right] &= E[ y^2 + \tilde{f}^2 -2y\tilde{f} ], \nonumber \\
&= E[ y^2 ] + E[ \tilde{f}^2 ] - 2E[ y \tilde{f} ], \nonumber \\
&= \text{Var}(y) +E[ y ]^2 + \text{Var}(\tilde{f}) + E[ \tilde{f} ]^2 - 2E[ y\tilde{f} ], \nonumber \\
&= \text{Var}(f) + \sigma^2 + \text{Var}(\tilde{f}) + E[ y ]^2 + E[ \tilde{f} ]^2 - 2E[ y\tilde{f} ], \nonumber \\
&= \text{Var}(f) + \sigma^2 + \text{Var}(\tilde{f}) + \big(E[ \tilde{f} ] - E[ f ]\big)^2 + 2E[ f ]E[ \tilde{f} ] - 2E[ y\tilde{f} ], \nonumber \\
&= \text{Var}(f) +\sigma^2 + \text{Var}(\tilde{f}) + E[\tilde{f} - f]^2 - 2\big(E[ y\tilde{f}] - E[ y ]E[ \tilde{f} ]\big), \nonumber \\
&= \text{Var}(f) +\sigma^2 + \text{Var}(\tilde{f}) + \text{Bias}(\tilde{f})^2 - 2\,\text{Cov}(y,\tilde{f}), \nonumber \\
&= \text{Var}(f) +\sigma^2 + \text{Var}(\tilde{f}) + \text{Bias}(\tilde{f})^2 - 2\,\text{Cov}(f,\tilde{f}),
\end{align}
where $\text{Bias}(\tilde{f}) = E[\tilde{f} - f]$. Now suppose, as mentioned above, that we want to know what the {\em expected test mean square error} $E[ (y(x_0) - \tilde{f}(x_0))^2 ]$ will be for a point $x_0 \in X \setminus \{x_1, x_2, \ldots, x_n\}$. This expectation is the average of the square error for all possible functions $\tilde{f}$ derived from all possible training sets when we perform the experiment described above repeatedly. This identifies $\tilde{f}(x_0)$ to a random variable.

In the finite case, let $\{(x_1^{(j)},y_1^{(j)}), (x_2^{(j)},y_2^{(j)}), \ldots, (x_n^{(j)},y_n^{(j)}) \}$ be the $j$-th training set that is obtained after performing the experiment $j$ times. Assume that we have $m$ training sets, then
\begin{equation} \label{eq: mean of a deterministic function}
E[ f(x_0) ] = \frac{1}{m} \sum_{j=1}^m f^{(j)}(x_0) = f(x_0),
\end{equation}
since $f^{(j)}(x_0) = f(x_0)$ because $f$ is deterministic, i.e., a regular function. From this,
\begin{equation} \label{eq: variance of a deterministic function}
\text{Var}(f(x_0)) = E\left[ \left(f(x_0) - E[f(x_0)]\right)^2 \right]= \frac{1}{m} \sum_{j=1}^m \left(f^{(j)}(x_0) - f(x_0)\right)^2 = 0.
\end{equation}
Similarly, we can prove that $\text{Cov}(f(x_0),\tilde{f}(x_0)) = 0$, and together with equations \eqref{eq: mean of a deterministic function} and \eqref{eq: variance of a deterministic function}, equation \eqref{eq: mean square error decomposition}  becomes
\begin{equation} \label{eq: mean square error at a point}
E\left[\big(y(x_0) - \tilde{f}(x_0)\big)^2\right] = \text{Var}\big(\tilde{f}(x_0)\big) + \text{Bias}\big(\tilde{f}(x_0)\big)^2 +\sigma^2,
\end{equation}
with
\begin{equation}
\text{Var}(\tilde{f}(x_0)) = \lim_{m \rightarrow \infty} \frac{1}{m} \sum_{j=1}^m \left(\tilde{f}^{(j)}(x_0) - \frac{1}{m}\sum_{i=1}^m \tilde{f}^{(i)}(x_0) \right)^2,
\end{equation}
and
\begin{equation}
\text{Bias}(\tilde{f}(x_0)) = \lim_{m \rightarrow \infty} \frac{1}{m} \sum_{j=1}^m \left( \tilde{f}^{(j)}(x_0) - f(x_0) \right).
\end{equation}

The overall expected test mean square error can be computed by averaging $E\big[ \big( y(x_0) - \tilde{f}(x_0) \big)^2 \big]$ over all possible values of $x_0$ in the test set \cite{JamWitHasTib2013}. 

Equation \eqref{eq: mean square error at a point} says that in order to minimize the expected test mean squared error, we must find a function $\tilde{f}$ that simultaneously yields low variance and low bias. Also note that since $\text{Var}(\tilde{f}(x_0)) \geq 0$ and $\text{Bias}(\tilde{f}(x_0))^2 \geq 0$, this error will always be at least as large as the variance $\sigma^2$ of the noise component $\epsilon$ in our model.

The variance speaks to how much $\tilde{f}$ would change if we changed our training data set, and bias refers to the error introduced by approximating a real-life problem by a much simpler model. Often these two quantities are mutually exclusive and therefore we speak of the bias-variance trade-off \cite{JamWitHasTib2013}.

In our recommender system model, the bias-variance trade-off is controlled by the number of features $k$, and the regularization factor $\lambda$. In figure \ref{fig: bias-variance trade-off} we show the error when reconstructing image I23 from noisy image I23\_06\_4 for values of $k$ equal to 1, 8, and 352 using a variety of values of $\lambda$ between 0 and 80. How to determine a {\em good} value of $k$? And how about $\lambda$?

\begin{figure}[htbp]
\centering
	\subfigure[]{\includegraphics[width=.485\textwidth]{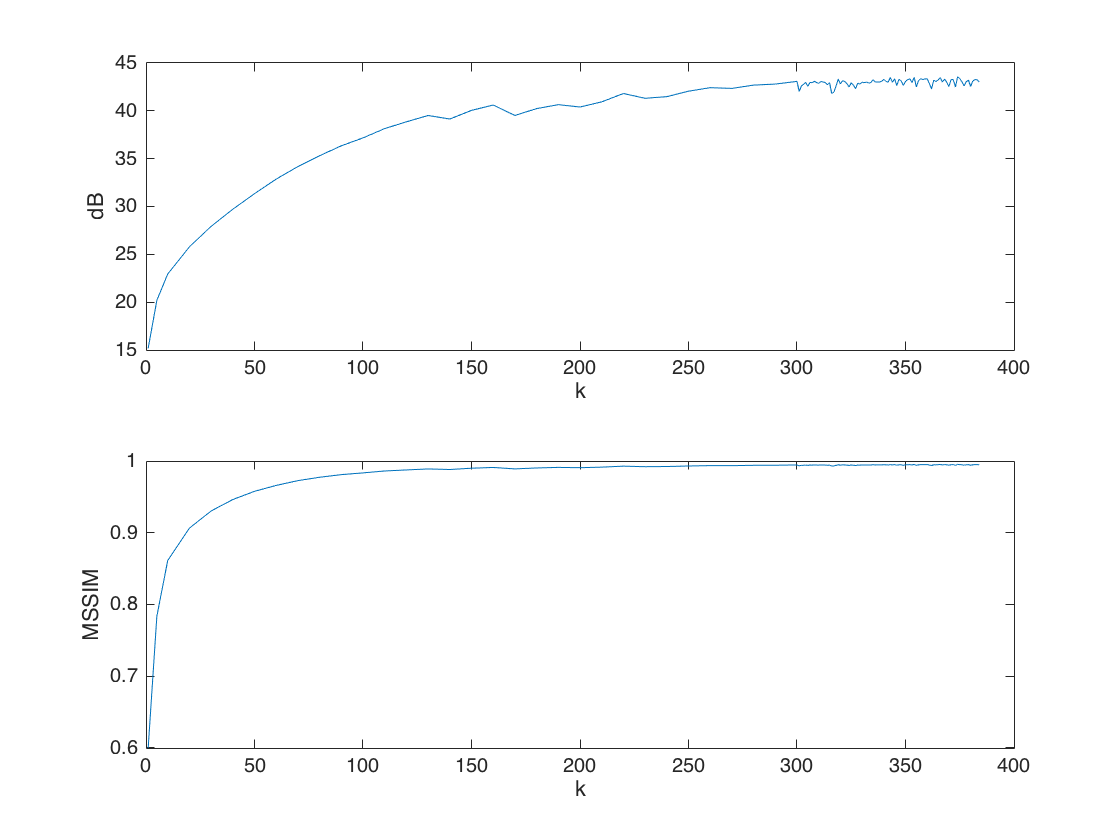} \label{fig: search of best k}}
	\subfigure[]{\includegraphics[width=.485\textwidth]{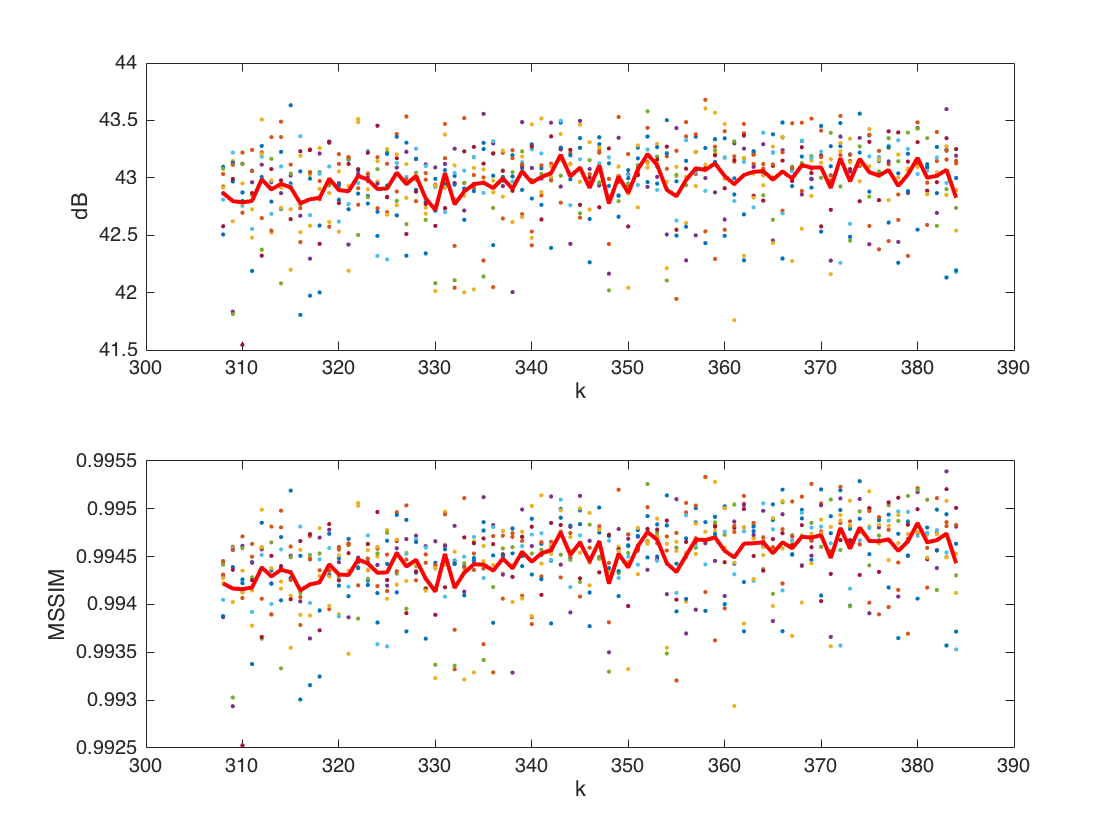} \label{fig: search of best k zoom}}
\caption{Search of an optimal number of features $k$ with the regularization factor $\lambda$ set to 0. Figure \subref{fig: search of best k} shows the PSNR in dB (top) and MSSIM values (bottom) versus the number of features $k$ for values of $k$ in $\{1, \ldots, 384\}$ when restoring image I23 from image I23\_06\_4. The image quality values were obtained with our recommender system random seed initialization set to 0 for all $k$ explored. The apparent smoothness for smaller values of $k$ is due to the fact that we sampled $k$ at larger intervals for that region of this parameter space. Figure \subref{fig: search of best k zoom} shows the results of the same exploration for $k=308, \ldots, 384$. The dots represent the values obtained for each of the same ten random seed initializations tried for each $k$. The red lines represent the averages of those ten experiments. For $k=352$ we obtain best average for PSNR, while $k=380$ is best for MSSIM.}
\label{fig: optimal k}
\end{figure}

\subsection{How to choose $k$ and $\lambda$ without a ground truth reference} \label{sec: choice of k and lambda}
For an $m \times n$ grayscale image we need $mn$ values to describe each pixel when no compression is used. For our recommender system model, we need to find two matrices $\mathbf{X}$ and $\Theta$ to propose a recommendation. If $\mathbf{X} \in \mathbb{R}^{k \times m}$ and $\Theta \in \mathbb{R}^{k \times n}$, see equation \eqref{eq: X and Theta}, then we need $km + kn$ numbers to represent those matrices. It is therefore reasonable to assume that we would need a value of $k$ such that $km + kn$ is at least as big as $mn$ to represent the image with fidelity. We consequently try to find the smallest $k \in \mathbb{N}$ such that $km + kn \geq mn$. Let $x \in \mathbb{R}$ and $m, n \in \mathbb{N}$, then
\begin{align}
xm + xn &= mn, \nonumber \\
 \Leftrightarrow \quad x &= \frac{mn}{m+n}, \nonumber
\end{align}
hence if $k_{min} \in \mathbb{N}$ is the smallest natural number such that $k_{min} \geq \frac{mn}{m+n}$, we would have at least as many numbers necessary to define our recommender system as there are numbers describing the image that we are trying to reconstruct. In our case, since all the images in our database are $384 \times 512$ color images, we have that $m = 384$ and $n = 3 \times 512 = 1536$, see section \ref{sec: experimental setup}. This gives a minimum $k_{min} = 308$ for a system that, with the criterion outlined above, will have enough variance to capture any of the pictures in our dataset.

\begin{figure}[htbp]
\centering
\includegraphics[width=1\textwidth]{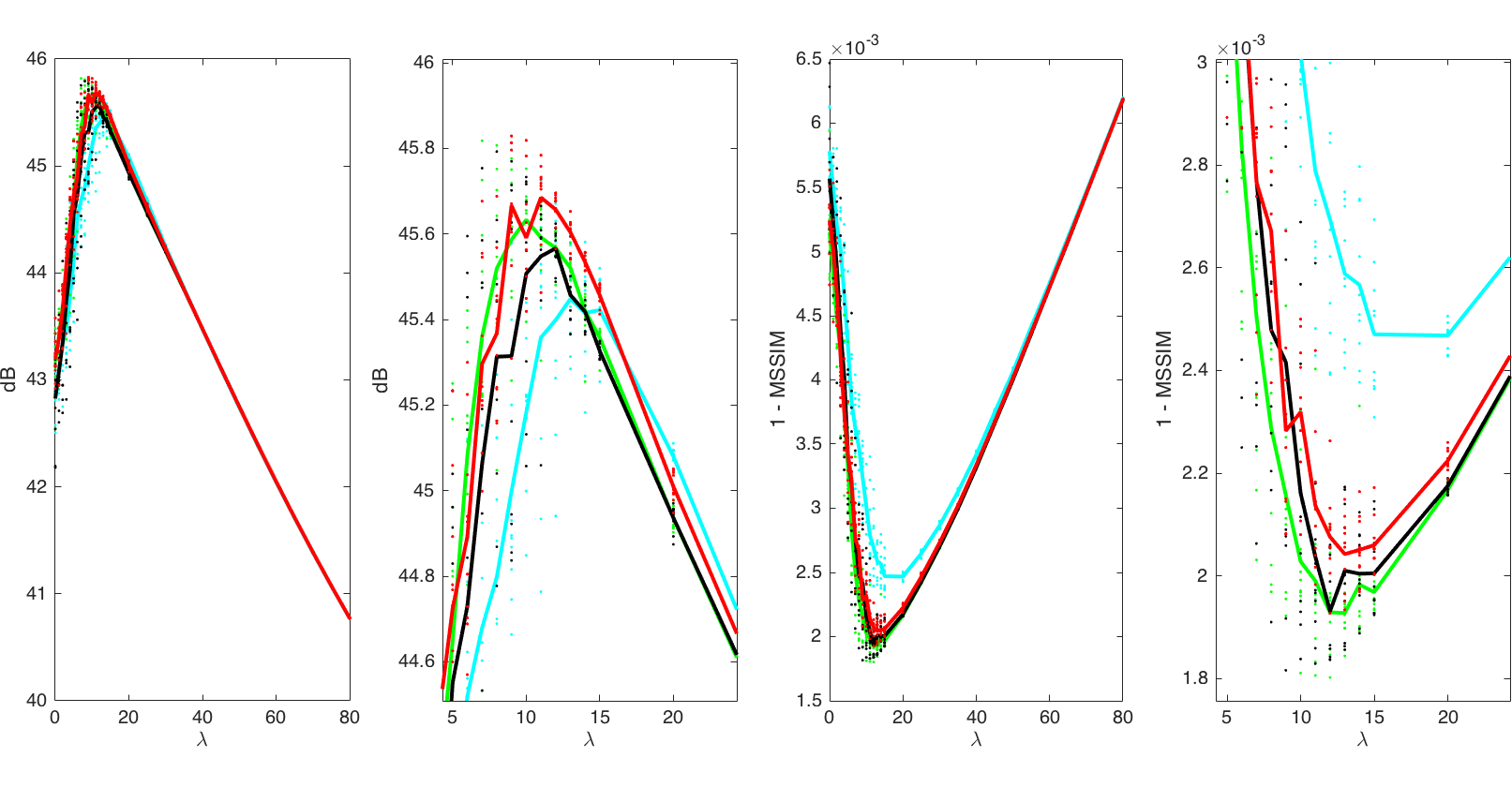}
\caption{Regularization and number of features optimization for image I23 for a noise ratio of $p=0.0678$. The solid lines correspond to the average of ten different random initializations for each of four values of $k$ explored in detail. The individual dots correspond to each of the ten runs from which the averages were computed. The cyan line and dots correspond to $k=308$, the red to $k=352$, the green to $k=380$, and the black to $k=384$ number of features, respectively. The two graphs on the left provide the peak signal-to-noise ratio in dB and the ones on the right correspond to the error as measured by $1-\text{MSSIM}$ (since two identical images will have an MSSIM value of 1), all as a function of the regularization factor $\lambda$. We found that $k=352$ with $\lambda=11$ optimized for PSNR, and $k=380$ with $\lambda = 13$ optimized for $1-\text{MSSIM}$.} \label{fig: optimal k and lambda experiments}
\end{figure}

On the other extreme, since $\frac{mn}{m+n} < \min\{m,n\}$ when $m,n \neq 0$, if we set $k_{max} = \min\{m,n\}$ we should also have enough features in our recommender system to capture the images in the database with fidelity. In our case $k_{max} = \min\{384,1536\} = 384$. Finally, through exploration, see figure \ref{fig: optimal k}, we find that $k=352$ is the number of features that maximizes the PSNR in our experiments with image I23 and noisy image I23\_06\_4 for a regularization factor of $\lambda = 0$, and $k=380$ maximizes MSSIM under the same circumstances.

To explore the impact of the regularization factor $\lambda$ we performed a numerical exploration for a range of values of $\lambda$ between 0 and 80. When reconstructing image I23 from image I23\_06\_4, we found that, on average, for $k=352$ with $\lambda=11$ we obtain the largest PSNR, and for $k=380$ with $\lambda = 13$ we get the smallest error as measured by $1 - \text{MSSIM}$. For $k=308, 384$ both image quality measures are not as good for their respective optimal values of $\lambda$. The experiments are summarized in figure \ref{fig: optimal k and lambda experiments}. They reflect the trade-off between bias and variance that we talked about since the parameter $\lambda$ controls, to a degree, for the variance in the model. For the experiments with all other images in our database we chose a regularization factor $\lambda = 11$ and we set $k=352$ for the number of features. We want to emphasize that we do not expect theses choices to produce optimal results for all images and all noise ratios $p$, but our experiments show that they produced very satisfactory results, as discussed in section \ref{sec: results}.

\begin{figure}[htbp]
\centering
	\subfigure[L-curves]{\includegraphics[width=.48\textwidth]{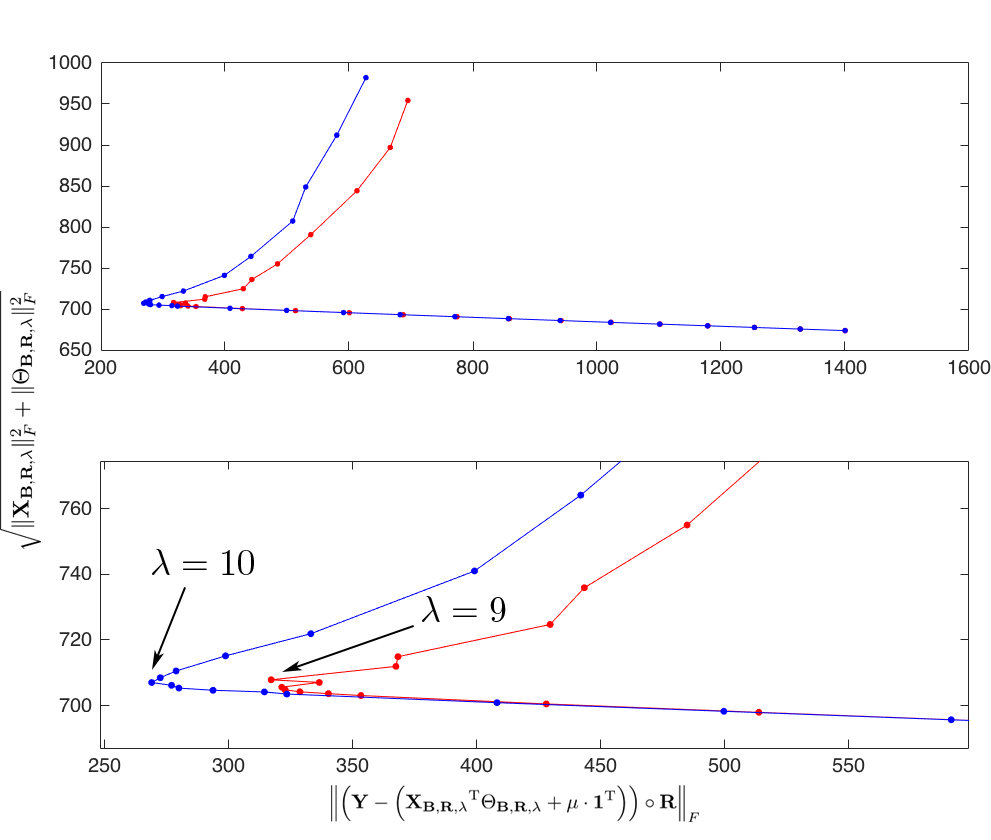} \label{fig: norm of the solution vs norm of the residue}}
	\subfigure[$\Sigma$-curves]{\includegraphics[width=.48\textwidth]{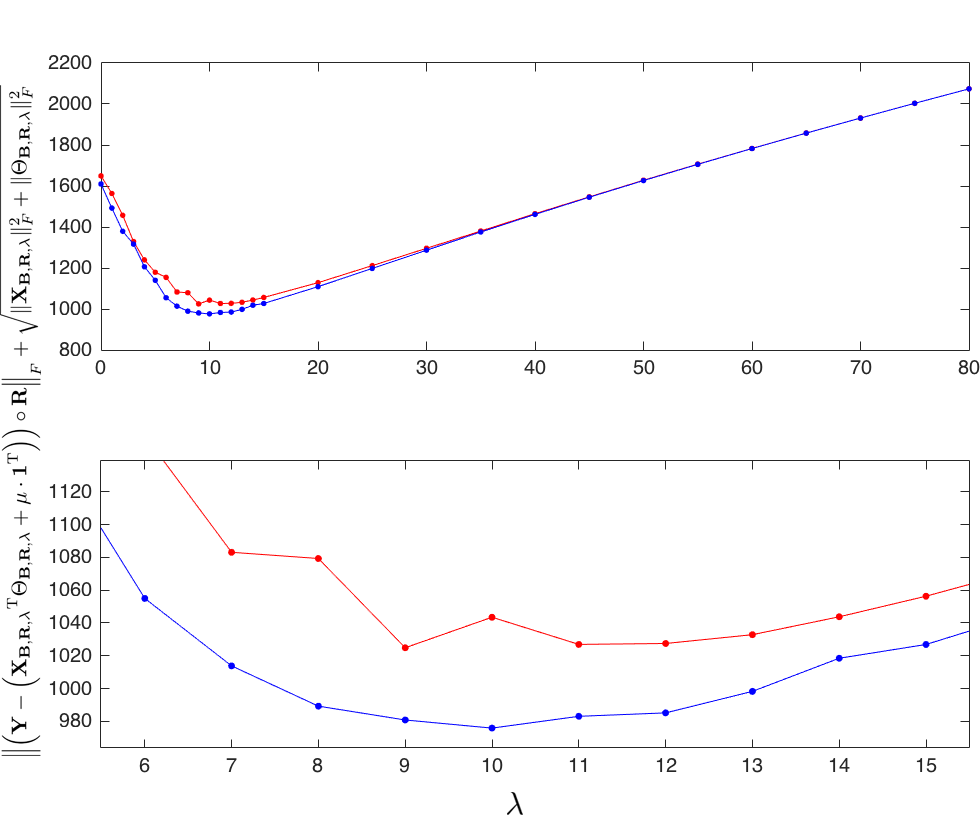} \label{fig: sum of norms vs lambda}}
	\caption{Search of an optimal regularization factor $\lambda$. In figure \subref{fig: norm of the solution vs norm of the residue} we plot the average solution norm versus the average residue norm computed from ten different random seeds when we reconstruct image I23 from I23\_06\_4, corresponding to a noise ratio of $p=0.0678$. Figure \subref{fig: sum of norms vs lambda} shows the sum of the quantities mentioned above versus the regularization factor $\lambda$. The red graphs correspond to $k=352$ number of features and the blue ones to $k=380$. The bottom graphs are detail versions of the top ones.}
\label{fig: L-curves}
\end{figure}

It is important to notice that for a given number of features $k$ we were able to find by exploration the optimal value for the regularization factor $\lambda$ because we counted with both the original and distorted versions of the particular image we chose to fine tune these parameters. But what if we didn't have that luxury, i.e., what if we only counted with the distorted version of the image to figure out a reasonable value for $\lambda$, as is the real world scenario? We propose in that case to use the L-curve method \cite{Hans1992,HanOle1993} to select the value of $\lambda$. In figure \ref{fig: norm of the solution vs norm of the residue} we plot the norm of the solution as given by,
\begin{equation} \label{eq: norm of solution}
\sqrt{\| \mathbf{X}_{\mathbf{B},\mathbf{R},\lambda} \|^2_F + \|\Theta_{\mathbf{B},\mathbf{R},\lambda} \|^2_F},
\end{equation}
versus the norm of the residue computed as,
\begin{equation} \label{eq: norm of residue}
\left\|\left(\mathbf{Y} - \left({\mathbf{X}_{\mathbf{B},\mathbf{R},\lambda}}^\text{T}\Theta_{\mathbf{B},\mathbf{R},\lambda} + \mu \cdot \mathbf{1}^\text{T}\right)\right) \circ \mathbf{R}\right\|_F,
\end{equation}
both averaged from ten different random seed reconstructions of image I23 from I23\_06\_4.

We observe that the value of the regularization factor $\lambda$ at the ``elbow" in the L-shaped curve is very close to the value of $\lambda$ for which we experimentally obtain the best reconstruction results, see table \ref{tab: parameter optimization}. That is, in the case of $k=352$ number of features, the L-shaped curve has an elbow at $\lambda = 9$; whereas for $k=380$, the elbow appears at $\lambda = 10$. These values of $\lambda$ are best seen when we plot the sum of the average residue and the average solution norms, and they correspond to the minima of what we call the $\Sigma$-curves in figure \ref{fig: sum of norms vs lambda}. What is remarkable of this approach is that we can find a very good approximation to the empirically obtained optimal value of $\lambda$ for a given $k$ via the L or $\Sigma$-curves without knowledge of the ground truth.

\begin{table}[htbp]
\centering
\caption{Parameter optimization for image I23\_06\_4}
	\begin{tabular}{l c c c c}
\toprule 
Method              & $k$ & $\lambda$ & PSNR (dB) & $1-\text{MSSIM}$ \\
\midrule 
\multirow{2}{*}{Empirical}	& 352	& 11	& 45.69	& \\
\cmidrule(r){2-5}
					& 380	& 13	& & 0.001928 \\
\midrule
\multirow{2}{*}{L-curve}	& 352	& 9	& 45.67	& \\
\cmidrule(r){2-5}
					& 380	& 10	& & 0.002028 \\
\midrule
\multirow{2}{*}{$\Sigma$-curve}	& 352	& 9 & 45.67 & \\
\cmidrule(r){2-5}
						& 380	& 10	& & 0.002028 \\
\bottomrule 
	\end{tabular}
  \label{tab: parameter optimization}
\end{table}

\subsection{Image decomposition and error: many choices} \label{sec: image decomposition}
Another topic to address is the image decomposition that we have proposed. In particular, the additive term $\mathbf{M}$ on the right hand side in the representation
\begin{equation}
\mathbf{D} = \mathbf{X}^\text{T}\Theta + \mathbf{M}.
\end{equation}
In our case we set $\mathbf{M} = \mu \cdot \mathbf{1}^\text{T} \in \mathbb{R}^{m \times 3n}$, with $m = 384$ and $n = 512$, where $\mu$ is the vector with entries $\mu_i = \frac{1}{\#\{j\, :\, r_{i,j}=1\}} \sum_{j\, :\, r_{i,j}=1} y_{i,j}$, and $\mathbf{1} \in \mathbb{R}^{3n}$ is the vector with all ones, see equation \eqref{eq: image restoration}. Matrix $\mathbf{M}$ is the term that provides normalization in our recommender system, see section \ref{sec: regularization and normalization}. This term averages the available pixel values in each row and proposes that average for any missing pixel in that row as a starting point, which is then further enhanced by the contribution coming from $\mathbf{X}^\text{T}\Theta$ for that particular pixel. This approach has the caveat that, in images, the value of a pixel may be more likely to be correlated with the values of nearby pixels, than with those that are farther away. This can be corrected by proposing different ways to build the normalizing matrix $\mathbf{M}$.

\begin{figure}[htbp]
\centering
	\subfigure[]{\includegraphics[width=.485\textwidth]{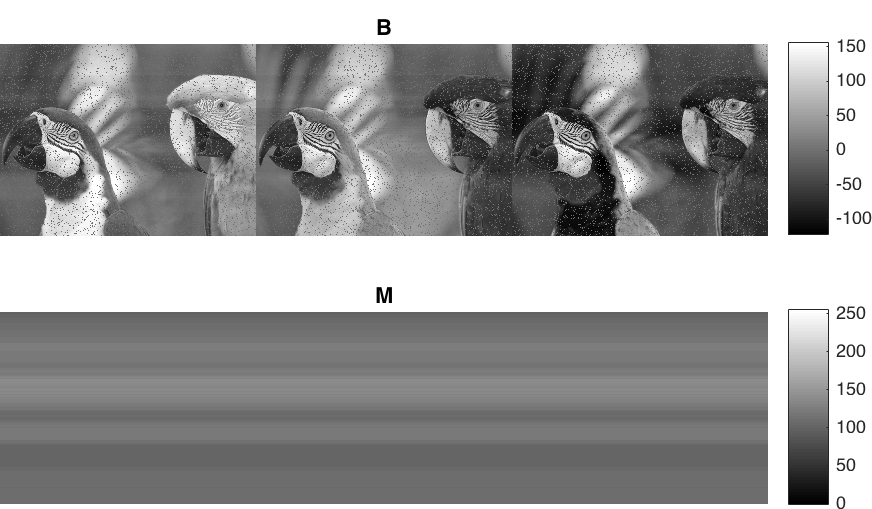} \label{fig: decomposition1}}
	\subfigure[]{\includegraphics[width=.485\textwidth]{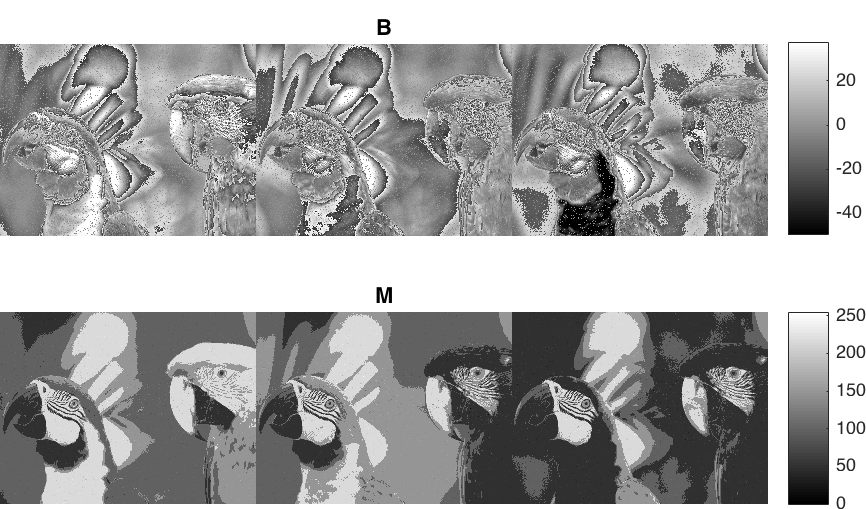} \label{fig: decomposition2}}
\caption{Two different decompositions of image I23\_06\_4. Adding both matrices in either column \subref{fig: decomposition1} or \subref{fig: decomposition2} gives the matrix representation of image I23\_06\_4, which is decomposed as the sum of a normalized term $\mathbf{B}$ and a mean term $\mathbf{M}$, see section \ref{sec: experimental setup}. Column \subref{fig: decomposition1} corresponds to the decomposition that we used in our experiments, column \subref{fig: decomposition2} corresponds to a mean matrix $\mathbf{M}$ that is produced via the $K$-means algorithm \cite{MacQ1967} with $K=4$. We use $\mathbf{B}$ in our algorithm to produce a decomposition $\mathbf{B} = \mathbf{X}^\text{T}\Theta$, which is further used to propose a reconstruction, see equation \eqref{eq: image restoration}.}
\label{fig: decompositions}
\end{figure}

One approach could be to assign a weight to the value of a contributing pixel that takes into account its distance to the reference pixel, for example; or to just average the pixel values of the pixels in the  {\em same class} that it belongs to, which could be defined in a variety of ways, such as belonging in the neighborhood of a $K$-means partition of the image, as another example, see figure \ref{fig: decomposition2}.

Finally, we want to mention how to address the slight asymmetry on how the rows and columns of the image are treated. We can transpose the image, process it in the same way the original image was processed, revert to the original orientation, and finally add and average both representations. We do this for our test image I23 and its noisy representation I23\_06\_4 and obtain that, over five different random initializations, we obtain an average improvement of $\approx 2.68$ dB in PSNR, and $7.2860 \times 10^{-4}$ units in MSSIM.

\section{Conclusions and future work} \label{sec: conclusions and future work}
We have proposed a collaborative filtering recommender system to restore images with impulse noise. This system assumes that we know the location of the pixels that have been corrupted by impulse noise and it uses this knowledge to build a recommendation for the values that those pixels should have. In the process we have proposed an image decomposition that consists in the product of two matrices plus the sum of a third one.

To test and calibrate this collaborative recommender system, we have used a well known image database. There are two parameters---$k$, the number of features; and $\lambda$, the regularization factor---which we have explored and optimized for our particular image data set to obtain the best image quality. We have measured the performance of our algorithm using two well known image quality metrics, PSNR and MSSIM. Our experiments show that this system performs very well at reconstructing the original images from their noisy counterparts provided we give a complete list of the distorted pixels in each of the originals.

We have provided a rationale on how to choose the parameters aforementioned, but this is an area of further exploration. Also, we have pointed out that the image decomposition proposed is not unique and we have hinted at other possible decompositions that fit our recommender system. Exploring these could lead to new insights, improvements, and applications of our algorithm.

\begin{figure}[htbp]
\centering
	\subfigure[Original]{\includegraphics[width=.3\textwidth]{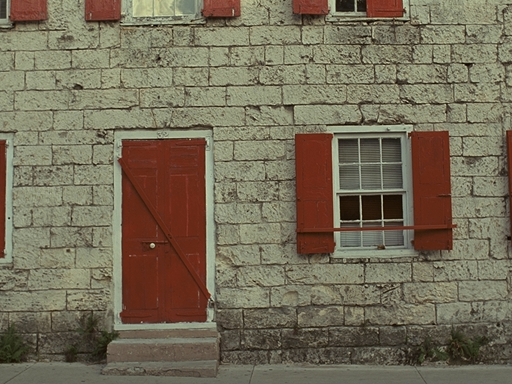} \label{fig: i01}}
	\subfigure[Mutilation]{\includegraphics[width=.3\textwidth]{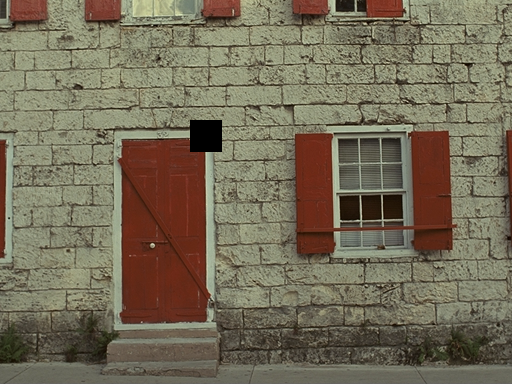} \label{fig: i01 square}}
	\subfigure[Reconstruction]{\includegraphics[width=.3\textwidth]{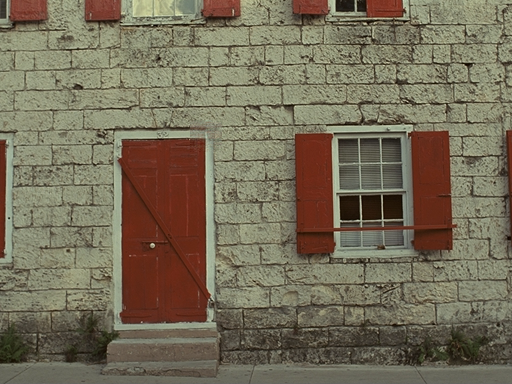} \label{fig: i01 inpainting}}
	\subfigure[Mutilation \cite{Wys2016}]{\includegraphics[width=.3\textwidth]{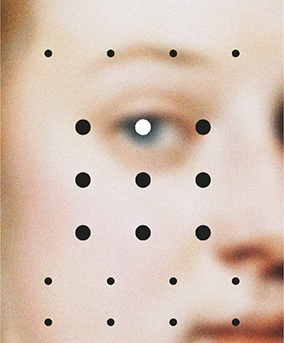} \label{fig: girl with dots}}
	\subfigure[Reconstruction]{\includegraphics[width=.3\textwidth]{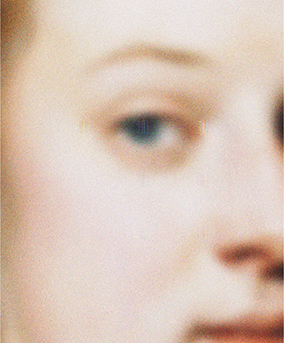} \label{fig: girl reconstruction}}
\caption{Image inpainting. Our collaborative filtering recommender system can be used to fill in missing portions of an image.}
\label{fig: inpainting}
\end{figure}

As described above, the bulk of our work focused on reconstructing images with isolated pixels deemed to be noisy, or missing. What happens when we consider missing segments of an image? Figures \ref{fig: i01}, \ref{fig: i01 square}, and \ref{fig: i01 inpainting} show the performance of our system when presented with an image that has been excised of a $32 \times 32$ square portion of its pixels. The reconstruction obtained compares remarkably well with the original and goes from a 29.21 dB mutilated image to a reconstruction at 41.81 dB. Figures \ref{fig: girl with dots} and \ref{fig: girl reconstruction} showcase an image whose original is not available that has been defaced with a grid of dots and for which our recommender system has provided a reconstruction that could be the original undisturbed image by visual inspection. These examples fall under the purview of the general techniques known as {\em inpainting} \cite{Ikeu2014,Shen2003}, and although the reconstructions shown are far from the current state-of-the-art in inpainting, the fact that our algorithm recovers the geometry and texture simultaneously without having been designed implicitly to solve this holy grail of inpainting makes exploring this application of our algorithm worthy of pursuit. One can envision as well applications of our work to superresolution.

I would like to thank the support of the leadership at the Institute for Physical Science and Technology, without which this work would not have been possible. I also want to personally thank John J.\ Benedetto, Director of the Norbert Wiener Center in the Mathematics Department at the University of Maryland, College Park; and Mike Dellomo, Associate Director and Advisor of the Masters in Telecommunications Program in the Department of Electrical and Computer Engineering also at the University of Maryland, College Park; for their feedback and valuable input in the writing of this manuscript. Finally, I want to acknowledge and thank the support of ARO Grants W911NF-15-1-0112 and W911NF-17-1-0014.

\bibliographystyle{plain}
\bibliography{JBbib}
 \end{document}